\documentclass[12pt]{amia}
\usepackage{multirow}
\usepackage{hyperref}
\usepackage{times}
\usepackage{xcolor}
\usepackage{adjustbox} 
\usepackage{tabularx} 

\begin{document}

\title{KU\_AIGEN\_ICL\_EDI@BC8 Track 3: Advancing Phenotype Named Entity Recognition and Normalization for Dysmorphology Physical Examination Reports}

\author{
Hajung Kim $^1$, Chanhwi Kim $^1$, Jiwoong Sohn $^1$, Tim Beck $^5$, Marek Rei $^6$, Sunkyu Kim $^3$,\\ T Ian Simpson $^4$, Joram M Posma $^2$, Antoine Lain $^{2,4}$, Mujeen Sung $^1$, Jaewoo Kang $^{1,3,*}$}

\institutes{
    $^1$ Department of Computer Science and Engineering, Korea University, $^2$ Department of Metabolism, Digestion and Reproduction, Imperial College London, $^3$ AIGEN Sciences, $^4$ Institute for Adaptive and Neural Computation, School of Informatics, University of Edinburgh, $^5$ School of Medicine, University of Nottingham, $^6$  Department of Computer Science, Imperial College London\\
}

\newcommand{\draftonly}[1]{#1}

\maketitle
\setlength{\parindent}{0.2in}
\section*{Abstract}

The objective of BioCreative8 Track 3 is to extract phenotypic key medical findings embedded within EHR texts and subsequently normalize these findings to their Human Phenotype Ontology (HPO) terms. However, the presence of diverse surface forms in phenotypic findings makes it challenging to accurately normalize them to the correct HPO terms. To address this challenge, we explored various models for named entity recognition and implemented data augmentation techniques such as synonym marginalization to enhance the normalization step. Our pipeline resulted in an exact extraction and normalization F1 score 2.6\% higher than the mean score of all submissions received in response to the challenge. Furthermore, in terms of the normalization F1 score, our approach surpassed the average performance by 1.9\%. These findings contribute to the advancement of automated medical data extraction and normalization techniques, showcasing potential pathways for future research and application in the biomedical domain.

\section*{Introduction}
Medical examinations describing physical dysmorphology involve documenting subtle variations observable in a patient's facial and bodily morphology. Despite the importance of these medical findings, they are often recorded as unstructured free text within electronic health records (EHR). This mode of recording imposes limitations on the utility of the data for extracting insightful inferences and conclusions. Thus, converting into standard vocabularies, such as the human phenotype ontology (HPO) \cite{hpoontology}, renders the data accessible for subsequent computational analysis. \par
Earlier works, such as Doc2HPO \cite{liu2019doc2hpo}, have leaned on dictionary-based matching. This approach utilizes a dictionary formulated from the HPO essential for identifying phenotypic terms. However, it falls short in dealing with unseen words that are absent from the preliminary dictionary. Recently, deep-learning based methods have shown improved performance compared to dictionary-based methods. Phenotagger \cite{luo2021phenotagger}, a hybrid method that combines both dictionary and deep-learning methods, and the PhenoBERT \cite{feng2022phenobert}, a deep-learning model, have achieved great improvements in the recognition of phenotype entities. Nevertheless, the performances of these works show limitations. All phenotype descriptions cannot be described continuously. Instead, one observation is scattered throughout a patient's record. For example, the observation ``long fingers and toes'' contains two phenotype terms, where `long' and `toes' are separated by an intervening word. Features like these make it difficult for computer systems to recognize and understand phenotype terms. Extracting information from these types of text records and then normalising them to HPO terms necessitates the use of sophisticated Natural Language Processing (NLP) techniques.

\section*{Materials and Methods}
The organizers provided the participants with two sets of data composed of observations extracted from dysmorphology physical examinations. The training set had 2,767 phenotype observations mapped to 1,716 unique consultations while the validation set had 734 phenotype observations mapped to 454 unique consultations. To solve this task we decomposed our pipeline into two parts: named entity recognition (NER) and named entity normalization (NEN).

\subsection*{A. Named Entity Recognition}
The first part of our pipeline aimed to identify HPO entities present in the input text using NER. \\
Looking at the 1,716 unique clinical consultations provided by the organizer for training, we identified four edge cases: observations with normal findings (i.e. normal lips), observations with no finding, discontinuous observations, and observations with only continuous findings. \par
We split the data 70\%/30\% taking into account the same proportion for every edge case to generate our training and validation sets. The validation set provided by the organizer was kept as such for testing. After analysing the data, we selected multiple models that showed state-of-the-art (SOTA) performance for biomedical NER (BioBERT \cite{lee2020biobert}, SciBERT \cite{beltagy2019scibert}), NER for HPO (PhenoTagger \cite{luo2021phenotagger}, PhenoBERT \cite{feng2022phenobert}) and discontinuous NER (TransE \cite{dai2020effective}, ChatGPT, W2NER \cite{li2022unified}). 
Following our optimization of each of these models we realized that ChatGPT and W2NER outperformed the other methods for both continuous and discontinuous cases and decided to focus on them only. \par
For our first NER model, we used the ChatGPT Finetuning API to finetune ChatGPT on our training dataset. Upon analyzing the characteristics of the dataset, we found that many sentences contained abbreviations, and some included mathematical symbols. The context for the abbreviations isn't included in the text, and the corpus trained on ChatGPT is not specialized enough to infer biomedical abbreviations, presenting a limitation. Therefore, we converted abbreviations into their full names. For sentences with statistical mathematical symbols, it is necessary to infer their meanings. We expanded these into full sentences; for example, ``HC $<$ 1\% for age'' was expanded to ``Head Circumference is below the 1st percentile for age.'' The extraction part using finetuned ChatGPT was conducted in two steps. The first step involved extracting all findings, regardless of whether they were key or normal. The second step consisted of classifying the extracted findings into their corresponding categories. \par
For our second NER model, we did a grid search to optimize the initial performance of W2NER \cite{li2022unified} by fine-tuning the hyper-parameters. The choice of pre-trained BERT model used in the first layer of the W2NER architecture had the biggest impact on model performance. We evaluated BioBERT, SciBERT, PubMedBERT and ClinicalBERT. ClinicalBERT showed the best performance with an improvement of 6.5\% F1 score over the worst performing model. The rest of the grid search resulted in training our model using 15 epochs, batch size of 8, learning rate of 0.001, BERT learning rate of 5e-5, and dropout of 0.3. W2NER was trained to extract only the key findings.

\subsection*{B. Named Entity Normalization}

To map entities that NER model found to correct HPO Term, we have to normalize those entities.
Our NEN pipeline uses embedding models, dictionary expansion, synonym marginalization, pre-finetuning, and additive synonym incorporation during marginalization. Our experiments showed that when combined these methods play a crucial role in enhancing entity normalization for biomedical text. We detail our approach and findings below.

To enhance entity normalization, we experimented with SapBERT \cite{liu2020self}. SapBERT is a pre-training scheme that self-aligns the representation space of biomedical entities. This is pre-trained with the UMLS 2020AA dataset. SapBERT achieved SOTA on six Medical Entity Linking datasets as reported in the original paper. Thus it led us to adopt SapBERT as our baseline embedding model. To make a generalisable representation and incorporate more synonyms of medical key findings, we used the HPO dictionary released on \href{https://github.com/obophenotype/human-phenotype-ontology/releases/tag/v2022-06-11}{(2022/06/22}. We flattened this dictionary and eliminated unobservable HPO terms, creating a more comprehensive dictionary for this challenge.

We employed the finetuning method, BioSyn \cite{sung2020biomedical} which used synonym marginalization to enhance our research. BioSyn maximizes the marginal likelihood of synonyms from the top candidates predicted by the model based on a dictionary. It iteratively updates candidates to include more challenging negative samples. For the training of BioSyn, we utilized the Biocreative training set. In an attempt to enhance the model's performance, we experimented with the exclusion of normal findings from the training set, but this adjustment did not result in significant performance improvements.
Recognizing that SapBERT was originally trained on UMLS data, which differs from the HPO term hierarchy, we undertook pre-fine tuning of SapBERT using the dictionary we constructed. This adaptation aimed to align SapBERT more closely with the HPO terminology. We call this pre-finetuned to HPO term model as PhenoSapBERT. In the process of synonym marginalization, we took an additional step by manually adding synonyms beyond what the model initially identified. We tried additive synonyms in k (k=1,3,5), with k=1 as the optimum.

\section*{Results}

For the challenge submission, we used the following NER models, \textbf{(1) Finetuned ChatGPT (FT-GPT)}, \textbf{(2) W2NER}, and \textbf{(3) Ensemble of (1), (2)}. each combined with, \textbf{BioSyn} for NEN. The final scores achieved on the test set by these models are shown in Table~\ref{tab:test_result}. The score presented in boldface means the highest score, and underlined score represents the second highest score. In Table~\ref{tab:test_result} NormOnly refers to the identification of HPO IDs for all key findings whether the spans are identified or not. The ExtNorm denotes the evaluation of the spans of key findings and their corresponding HPO term IDs, applied for each exact match and overlapping match of spans. We achieved 2.6\% higher in ExactExtNormF1 than the mean score of all submissions of the challenge. Furthermore, in terms of the NormOnlyF1, our approach surpassed the average performance by 1.9\%. 

\begin{table*}[h]
\centering
\footnotesize
\resizebox{\columnwidth}{!}{%
\renewcommand{\arraystretch}{1}
\begin{tabular}{|l|ccc|ccc|ccc|}
\hline
\multirow{2}{*}{\textbf{Method}} & \multicolumn{3}{c|}{\textbf{NormOnly}} &\multicolumn{3}{c|}{\textbf{ExactExtNorm}} & \multicolumn{3}{c|}{\textbf{OverExtNorm}} \\  \cline{2-10} & \textbf{Precision} & \textbf{Recall} & \textbf{F1} &  \textbf{Precision} & \textbf{Recall} & \textbf{F1} & \textbf{Precision} & \textbf{Recall} & \textbf{F1}  \\ \hline
\textbf{FT-GPT}              & \(\underline{73.47}\) & \(\underline{75.52}\) &\(\textbf{74.48}\) & \(\underline{70.28}\)&\(64.47\)&\(67.25\) & \(\underline{73.39}\) & \(\underline{75.20}\) &\(\textbf{74.28}\) \\
\textbf{W2NER}                  & \(\textbf{75.70}\) & \(73.29\) &\(\underline{74.47}\) & \(\textbf{73.64}\)&\(\underline{65.74}\)&\(\textbf{69.47}\) & \(\textbf{75.62}\) & \(72.97\) &\(\underline{74.27}\) \\
\textbf{Ensemble}                 & \(71.76\) & \(\textbf{76.95}\) &\(74.26\) & \(69.35\)&\(\textbf{68.52}\)&\(\underline{68.93}\) & \(71.69\) & \(\textbf{76.71}\) &\(74.12\) \\ \hline

\end{tabular}
}
\caption{Official score of our pipeline for the challenge obtained from the organizer.}
\label{tab:test_result}
\end{table*}

Table~\ref{tab:normalization} shows the ablation study of each NEN step using gold NER annotations. 
Not applying additive synonyms results in a 0.16\% decrease in the F1 score, while not applying pre-finetuning leads to a further drop of 1.39\% in the F1 score. Furthermore, the absence of fine-tuning results in a further decrease of 6.20\% in the F1 score. This underscores the significance of fine-tuning for enhancing entity normalization performance.

\begin{table*}[h]
\centering
\footnotesize
\begin{tabular}{|l|ccc|}
\hline
\multirow{2}{*}{\textbf{Method}} & \multicolumn{3}{c|}{\textbf{NormOnly}} \\  \cline{2-4} & \textbf{Precision} & \textbf{Recall} & \textbf{F1}  \\ \hline
\textbf{PhenoSapBERT + BioSyn (Ours)}  &  \textbf{80.12} & \textbf{90.28} & \textbf{84.90} \\
\textbf{- Additive Synonyms}      & 79.97 & 90.12 & 84.74 \\
\textbf{\quad- Pre-finetuning}        & 78.65 & 88.63 & 83.35 \\
\textbf{\qquad- Fine-tuning}       & 72.81 & 82.04 & 77.15 \\ \hline

\end{tabular}

\caption{
Ablation study of each NEN step on the BioCreative8 Task 3 validation dataset.
Not applying pre-finetuning denotes using SapBERT as an initial model instead of using PhenoSapBERT. Not applying fine-tuning refers to SapBERT without fine-tuning it with BioSyn.}
\label{tab:normalization}

\end{table*}

\section*{Conclusion}
We tested various methods on BioCreative8 Task 3 test set with unstructured free text derived from EHR. Our pipeline first extracts key findings and then normalizes concepts to HPO term IDs. For the extraction of key findings within EHR text, particularly involving disjoint key findings, the model designed to handle discontinuous concepts demonstrated robust performance. The enhancement of entity representation in BioSyn facilitates the mapping of key findings to HPO term IDs. These insights advance the field of automated extraction and normalization of medical data for future research and applications within the biomedical domain.

\section*{Funding}
This research was supported by (1) National Research Foundation of Korea (NRF2023R1A2C3004176, RS-2023-00262002), (2) the MSIT (Ministry of Science and ICT), Korea, under the ICT Creative Consilience program (IITP-2023-2020-0-01819) supervised by the IITP (Institute for Information \& communications Technology Planning \& Evaluation), (3) a grant of the Korea Health Technology R\&D Project through the Korea Health Industry Development Institute (KHIDI), funded by the Ministry of Health \& Welfare, Republic of Korea (grant number: HR20C0021(3)), (4) School of Informatics, University of Edinburgh and (5) The CoDiet project is funded by the European Union under Horizon Europe grant number 101084642 and supported by UK Research and Innovation (UKRI) under the UK government’s Horizon Europe funding guarantee [grant number 101084642].										

\makeatletter
\renewcommand{\@biblabel}[1]{\hfill #1.}
\makeatother

\bibliographystyle{vancouver}
\bibliography{amia}  

\begin{thebibliography}{10}

\bibitem{hpoontology}
Köhler S, et~al.
\newblock The Human Phenotype Ontology in 2021.
\newblock Nucleic Acids Research. 2021;49(D1):D1207-17.

\bibitem{liu2019doc2hpo}
Liu C, Peres~Kury FS, Li Z, Ta C, Wang K, Weng C.
\newblock Doc2Hpo: a web application for efficient and accurate HPO concept curation.
\newblock Nucleic acids research. 2019;47(W1):W566-70.

\bibitem{luo2021phenotagger}
Luo L, Yan S, Lai PT, Veltri D, Oler A, Xirasagar S, et~al.
\newblock PhenoTagger: a hybrid method for phenotype concept recognition using human phenotype ontology.
\newblock Bioinformatics. 2021;37(13):1884-90.

\bibitem{feng2022phenobert}
Feng Y, Qi L, Tian W.
\newblock PhenoBERT: a combined deep learning method for automated recognition of human phenotype ontology.
\newblock IEEE/ACM Transactions on Computational Biology and Bioinformatics. 2022;20(2):1269-77.

\bibitem{lee2020biobert}
Lee J, Yoon W, Kim S, Kim D, Kim S, So CH, et~al.
\newblock BioBERT: a pre-trained biomedical language representation model for biomedical text mining.
\newblock Bioinformatics. 2020;36(4):1234-40.

\bibitem{beltagy2019scibert}
Beltagy I, Lo K, Cohan A.
\newblock SciBERT: A pretrained language model for scientific text.
\newblock arXiv preprint arXiv:190310676. 2019.

\bibitem{dai2020effective}
Dai X, Karimi S, Hachey B, Paris C.
\newblock An effective transition-based model for discontinuous NER.
\newblock arXiv preprint arXiv:200413454. 2020.

\bibitem{li2022unified}
Li J, Fei H, Liu J, Wu S, Zhang M, Teng C, et~al.
\newblock Unified named entity recognition as word-word relation classification.
\newblock In: Proceedings of the AAAI Conference on Artificial Intelligence. vol.~36; 2022. p. 10965-73.

\bibitem{liu2020self}
Liu F, Shareghi E, Meng Z, Basaldella M, Collier N.
\newblock Self-alignment pretraining for biomedical entity representations.
\newblock arXiv preprint arXiv:201011784. 2020.

\bibitem{sung2020biomedical}
Sung M, Jeon H, Lee J, Kang J.
\newblock Biomedical entity representations with synonym marginalization.
\newblock arXiv preprint arXiv:200500239. 2020.

\end{thebibliography}

\end{document}